\documentclass[10pt,twocolumn,letterpaper]{article} 

\usepackage{float}
\usepackage{avss}
\usepackage{times}
\usepackage{epsfig}
\usepackage{graphicx}
\usepackage{amsmath}
\usepackage{amssymb}

\usepackage{hyperref}
\usepackage[noabbrev,capitalize,nameinlink]{cleveref}
\usepackage{multirow}

\usepackage{multibib}
\newcites{supp}{Supplementary References}

\avssfinalcopy 

\def\avssPaperID{17} 

\ifavssfinal\fi
\begin{document}

\title{Enhancing Zero-Shot Pedestrian Attribute Recognition with Synthetic Data Generation: A Comparative Study with Image-To-Image Diffusion Models
}

\author{
Pablo Ayuso-Albizu, Juan C. SanMiguel, Pablo Carballeira\\
\textit{Universidad Autónoma de Madrid, Madrid, Spain}\\
\tt\small pablo.ayuso@estudiante.uam.es, juancarlos.sanmiguel@uam.es, pablo.carballeira@uam.es}

\maketitle

\begin{abstract}
   Pedestrian Attribute Recognition (PAR) involves identifying various human attributes from images with applications in intelligent monitoring systems. The scarcity of large-scale annotated datasets hinders the generalization of PAR models, specially in complex scenarios involving occlusions, varying poses, and diverse environments. Recent advances in diffusion models have shown promise for generating diverse and realistic synthetic images, allowing to expand the size and variability of training data. However, the potential of diffusion-based data expansion for generating PAR-like images remains underexplored. Such expansion may enhance the robustness and adaptability of PAR models in real-world scenarios. This paper investigates the effectiveness of diffusion models in generating synthetic pedestrian images tailored to PAR tasks. We identify key parameters of \textit{img2img} diffusion-based data expansion —including text prompts, image properties, and the latest enhancements in diffusion-based data augmentation—and examine their impact on the quality of generated images for PAR. Furthermore, we employ the best-performing expansion approach to generate synthetic images for training PAR models, by enriching the zero-shot datasets. Experimental results show that prompt alignment and image properties are critical factors in image generation, with optimal selection leading to a 4.5\% improvement in PAR recognition performance.
\end{abstract}

\section{Introduction}

Pedestrian Attribute Recognition (PAR) focuses on analyzing full-body human data within video monitoring applications such as crowd analysis, person search and tracking. The increasing volume of human-centric data has driven the demand for automated PAR solutions. However, real-world datasets often present challenges like data redundancy and limited diversity, which can lead to inefficiencies in model training and reduced generalization capabilities \cite{har_survey}.

To address these challenges, data augmentation becomes essential to train in deep learning by altering the style or appearance of existing images. However, it operates within the constraints of a fixed dataset, without introducing entirely new samples. In contrast, \textit{dataset expansion based on image synthesis} leverages generative models to create novel synthetic images that capture the training data distribution, enhancing the original dataset diversity \cite{gans_avss} \cite{dm_survey}. These methods may increase training performance by maintaining semantic content while increasing variability in image layout and style, tackling the redundancy or poor diversity of the datasets. Diffusion models, a type of generative model, can also synthesize highly realistic and diverse images, giving novel images, with high quality and training stability \cite{is_synthetic_data}.

Recent studies on data augmentation through diffusion-based image synthesis \cite{method_alia} \cite{method_gif} \cite{method_dafusion} \cite{method_dalda} primarily focus on generating diverse classes. However, the variations between inter-class are significantly larger than those observed in pedestrian attributes. Consequently, there has been limited research on the generation of synthetic PAR-like images.

In this paper, we present a comparative study on the generation of PAR images. This study examines the impact of key variables such as prompts, image properties and major data expansion modifications, within the context of the diffusion pipeline \cite{stable_diffusion}. Additionally, we evaluated the impact of training  a PAR method with real images and synthetic images, highlighting their limitations and potential areas for improvement for zero-shot datasets PARzs. Our findings indicate that training with synthetic data improves overall performance and enhances the robustness of PAR models.

The contributions of this paper are as follows:
\begin{itemize}
    \item We \textbf{identified} key components within diffusion-based data expansion like prompt and image properties.
    \item We \textbf{explored} the impact of the selected parameters, guidance factors in diffusion and state-of-the-art strategies on the quality of the generated images for PARzs. 
    \item We \textbf{evaluated} how synthetic data could benefit the training of competitive PAR methods for PARzs.
\end{itemize}

\section{Related work}

\subsection{Pedestrian Attribute Recognition}
\label{subsec:par}
PAR aims at identifying semantic attributes of pedestrians from images. These attributes typically include characteristics such as age group (e.g., adult or child), gender, clothing style, carried objects (e.g., bag or umbrella), and other visual traits. PAR presents significant challenges due to multiple viewpoints, occlusions, unbalanced attribute distribution, low resolution, illumination variations, and image blur \cite{par_survey}. Over the years, numerous datasets have been proposed for training PAR methods, each of them varying in scale, complexity, and the number of attributes. PETA \cite{peta_dataset} presented a 19K images with 65 attributes. RAP \cite{rap_dataset} proposed a higher number of annotated images with 41K images with 69 attributes. Moreover, RAPv2 \cite{rap2_dataset} extended RAP with 85K images and 72 labeled attributes. Many of these datasets contain overlapping pedestrian identities between the training and test sets, so the evaluation becomes biased. Consequently, zero-shot PAR datasets (PARzs) have emerged as a solution for this issue, such as the largest dataset PA100K \cite{pa100k_dataset} with 100K images and 26 attributes, PETAzs (train: 19K, val: 11K, test: 3K) and RAPzs (train: 17K, val: 4.6K, test: 5K) \cite{rethinking_2021}. PARzs datasets are more challenging than non-ZS ones because models must generalize well enough to classify identities not found during training. PARzs datasets demonstrate that PAR methods are sensitive to the identities since the mA performance decreases in PETAzs (17\%) and RAPzs (9\%) \cite{rethinking_2021}. 

\subsection{Synthetic data generation}
\label{subsec:syndatagen}
Synthetic data generation is mostly based on Generative Adversarial Networks (GANs) \cite{gans_goodfellow} or Diffusion Models (DMs). GANs have demonstrated remarkable success in tasks such as image synthesis and style transfer \cite{gans_survey}. Advances in DMs have proven effective in creating realistic images that preserve the original data distribution, allowing to enrich training data \cite{dm_survey}. Despite of GANs effectiveness in image syntheis, DMs outperformed them \cite{diffusion_vs_gans}.

Recently, Latent Diffusion Models (LDMs) \cite{stable_diffusion} have gained attention for cost-efficient and high-quality image generation while working in a low-dimensional latent space. LDMs offer two main pipelines \cite{stable_diffusion}: text-to-image (\textit{txt2img}) and image-to-image (\textit{img2img}). While \textit{txt2img} provides flexibility for image editing \cite{diffusion_txt2img_sine}, it often lacks fine control over details, leading to inconsistencies in pedestrian attributes. In contrast, \textit{img2img} ensures structural coherence by modifying a reference image while preserving realistic human features \cite{diffusion_img2img_cyclenet}. Therefore, inspired by the success of Real Guidance \cite{is_synthetic_data}, we explored \textit{img2img} LDMs due to their lower computational cost and ability to generate high-quality images with consistent pose and appearance.

\subsection{Synthetic data to enhance training}
\label{subsec:syndatatraining}
Dataset expansion strategies help address challenges of small-scale datasets by generating additional synthetic images. The objective is improving the diversity of training samples and boosting machine learning models' robustness and generalization. Expansion of training datasets by synthetic images generated using GANs have improved the performance in tasks like person re-identification (ReID) \cite{gans_reid} and PAR \cite{gans_avss}. Focused on image class augmentation by LDMs, we have examples as prompt variations like ALIA \cite{method_alia} and DALDA \cite{method_dalda} that introduced variability to the dataset expanded. DALDA also proposed leverage strength hyperparameter by CLIPScore. We have others works focused on adding variations to the LDM pipeline like GIF \cite{method_gif}. Moreover, DAFUSION \cite{method_dafusion} uses Textual Inversion \cite{textual_inver} to learn a token for generating conditional images. Other works, focused in the ReID task, have demonstrated that diffusion synthesis images could improve the models \cite{reid_diffusion_syn}. Followed by that idea, there are some works that includes pose estimations to improve the image synthesis \cite{reid_diffusion_syn_pose}. However, those works focused on generating variations from existing identities in the training dataset rather than generating entirely new data to increase diversity.


\begin{figure*}[t]
\centering
\includegraphics[scale=0.6]{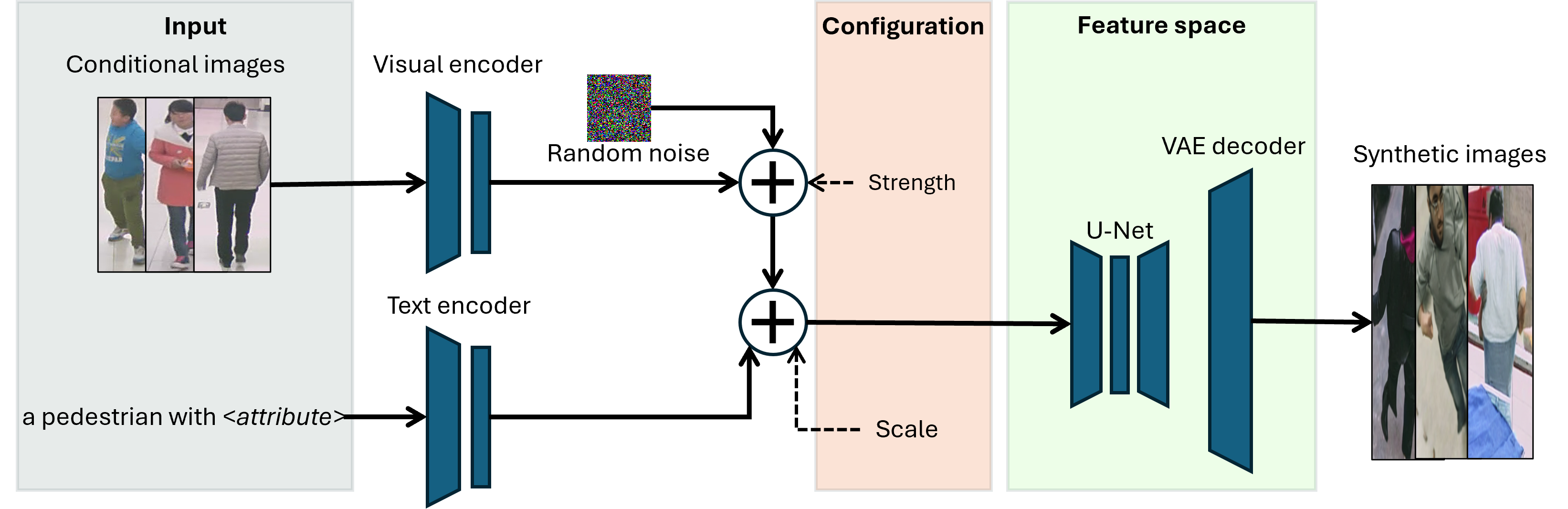}
\caption{Proposed conceptual framework to study image-guided diffusion models for synthetic data augmentation in Pedestrian Attribute Recognition, and the key components identified: inputs (gray box), configuration (magenta box) and feature space (green box).}
\label{fig:schemaofexperiments}
\vspace{-2mm}
\end{figure*}

\section{Image-guided diffusion model components}
\label{sec:canonicalsystem}

In this section, we describe the key components of our proposal for synthetic data augmentation using image-guided diffusion pipelines. Our proposal serves as a conceptual framework to study related methods for synthetic data generation of PAR-like images. As illustrated in \cref{fig:schemaofexperiments}, we consider three main components: input, configuration and feature space. 
The first key component is the \textbf{input}, being composed by \textit{prompts} and \textit{images} that allow conditioning the outputs.
The second key component is the \textbf{configuration} of hyperparameters, where \textit{scale} and \textit{strength} are identified as critical. The \textit{scale} determines the influence of the text prompt on guiding the model's output. The \textit{strength} controls the extent to which the conditional image is altered by adding noise. Normally, this parameter is fixed, but DALDA \cite{method_dalda} proposed to calculate the strength based on the distance between the image and the prompt.
The third component is the \textbf{feature space}, where the inputs are encoded into vectors. Then, random noise is added to the image vector depending on the strength. Later, the text vector is combined with the image vector, relying on the scale. Once the final encoded vector is created, the U-Net \cite{unet} removes the added noise. After noise removal, a Variational AutoEncoder upscales the vector to generate a new sample. Our conceptual framework allows us to explore advanced techniques such as the textual inversion of DAFUSION \cite{method_dafusion} and the latent modification of GIF \cite{method_gif}.

\section{Methodology}
\label{sec:methodology}

\subsection{Experiment design}
\label{subsec:expdesign}
\begin{figure}[t]
\centering
\includegraphics[scale=0.35]{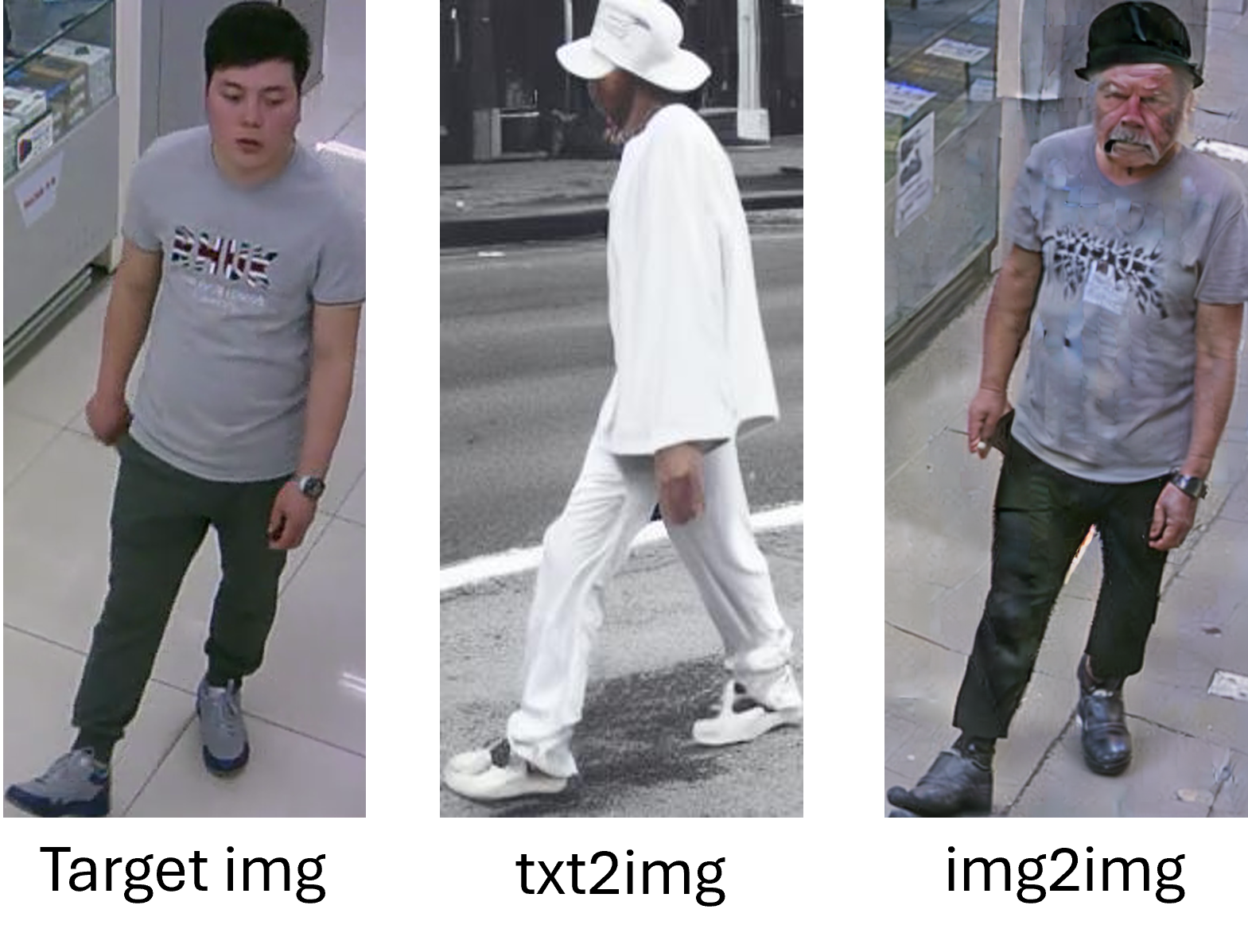}
\caption{Visual comparison between Text-to-Image and Image-to-Image diffusion-based pipelines for PAR-like image generation. Prompt used: \textit{A photo of a pedestrian with black hat}.}
\label{fig:visualcomtxt2imgimg2img}
\vspace{-2mm}
\end{figure}

We explore how \textit{img2img} diffusion pipelines could reduce the domain gap between generated and real images for PAR. In \cref{fig:visualcomtxt2imgimg2img} we observe how a conditional image could bring closer the generated image. Based on the components explained in \cref{sec:canonicalsystem}, we defined experiments for prompts, image properties, techniques modifying components and dataset expansion. After these experiments, we validated our conclusions by training with an expanded dataset. For more experiments, adapted prompts, examples of prompt and more generated images from each experiments check the supplementary material.

Moreover, we considered three setups for the \textbf{configuration block} mentioned in \cref{sec:canonicalsystem}. Since the \textit{strength} parameter ranges from 0.0 (low alteration) to 1.0 (high alteration) and the \textit{scale} parameter does not have fixed range, we proposed the following configurations:  high strength-low scale (0.6-3, HiSt LoSc), high strength-high scale (0.6-15, HiSt HiSc), low strength-low scale (0.2-3, LoSt LoSc).

For the \textbf{prompt} experiments, we selected the MALS \cite{mals_dataset} dataset because it contains high-resolution images (e.g., 575x230). This dataset allowed us to avoid working with low resolution (as seen in RAP or RAPzs \cite{rethinking_2021} images), ensuring a controlled experiment. For this study, we selected the 'c\_g\_a\_0' split of the MALS dataset. For \textbf{ image properties}, we examined blurriness and the surrounding spatial context of pedestrians. We aimed to assess their impact by introducing blurring or removing context, as RAPzs images already lack clarity and context. To achieve this, we used the MALS dataset for blurring and the TJU-DHD dataset \cite{dhd_dataset} for context removal. For TJU-DHD dataset we selected a subset from the pedestrian (campus) category.

Once we have explored the impact of the inputs, we compared \textbf{recent techniques} that introduce changes in the blocks explored: in the input (Textual Inversion \cite{textual_inver} by DAFUSION \cite{method_dafusion}), in the configuration (DALDA \cite{method_dalda}), and in the feature space (GIF \cite{method_gif}). For this type of experiments, we selected the RAPzs since we want to select the best technique to train a PAR method with synthetic data. Thus, between RAPzs and PETAzs, we decided RAPzs since has more diversity since has more images and attributes.

\subsection{Evaluation metrics}
To assess the quality of data generation, we used the Fréchet Inception Distance (FID) \cite{FID_metric}, which measures the distribution distance between two image sets by means of a pre-trained Inception network \cite{dm_survey}. 
For each experiment, 100 conditional real images were randomly selected, each generating one synthetic output. 
Then, we computed the FID between the generated images and the full dataset. As a reference, FID scores for random 100-image subsets are 105.24 (MALS), 130.44 (DHD), and 119.48 (RAPzs).
To evaluate PAR, we used the mean accuracy (mA) at attribute level \cite{rethinking_2021}, as indicator of attribute performance.

\section{Experimental results}
\label{sec:expresults}
\subsection{Setup}

For prompt-based and image-based experiments, we used Stable Diffusion v1.4 \cite{stable_diffusion}, while technique-based experiments relied on authors' available code. All experiments were conducted on a single NVIDIA A40 GPU.

\subsection{Results for varying prompts}
\label{subsec:img2imgprompt}

\begin{table}[]
\centering
\resizebox{0.85\columnwidth}{!}{
\begin{tabular}{|cc|ccc|}
\hline
\multicolumn{2}{|c|}{\textbf{Prompt used}} & \multicolumn{3}{c|}{\textbf{FID score per configuration}} \\ \hline
\multicolumn{1}{|c|}{\textbf{Generated by}} & Name & \multicolumn{1}{c|}{HiStr HiSc} & \multicolumn{1}{c|}{HiStr LoSc} & LoStr LoSc \\ \hline
\multicolumn{1}{|c|}{\multirow{2}{*}{Handcrafted}} & Baseline & \multicolumn{1}{c|}{208} & \multicolumn{1}{c|}{232} & 208 \\ \cline{2-5} 
\multicolumn{1}{|c|}{} & Integration \cite{prompt_integrating} & \multicolumn{1}{c|}{205} & \multicolumn{1}{c|}{221} & 136 \\ \hline
\multicolumn{1}{|c|}{\multirow{2}{*}{LLM}} & ALIA \cite{method_alia} & \multicolumn{1}{c|}{191} & \multicolumn{1}{c|}{203} & 136 \\ \cline{2-5} 
\multicolumn{1}{|c|}{} & DALDA \cite{method_dalda} & \multicolumn{1}{c|}{188} & \multicolumn{1}{c|}{205} & 136 \\ \hline
\multicolumn{1}{|c|}{\multirow{2}{*}{MALS caption}} & Not Aligned \cite{mals_dataset} & \multicolumn{1}{c|}{198} & \multicolumn{1}{c|}{214} & 138 \\ \cline{2-5} 
\multicolumn{1}{|c|}{} & Aligned \cite{mals_dataset} & \multicolumn{1}{c|}{\textbf{179}} & \multicolumn{1}{c|}{\textbf{185}} & \textbf{135} \\ \hline
\end{tabular}
}
\caption{Comparative results for prompt-based experiments to generate PAR-like images. FID compared against conditional images from MALS. Best results in bold.}
    \label{fig:fidcomprompts}
    \vspace{-3mm}
\end{table}

We investigated the impact of \textbf{prompt variations} on image generation, testing three types: handcrafted, LLM-generated and dataset-derived captions. The baseline handcrafted prompt is structured as \textit{A photo of a pedestrian wearing [color] [clothes]}, with color options $\{$white, black, red, blue, yellow, orange$\}$, and clothing options $\{$sweater, t-shirt, dress, jeans, hat, hair$\}$. The second handcrafted option follows the approach in \cite{prompt_integrating}. The first LLM-generated prompt is from DALDA \cite{method_dalda}, which feeds the LLM with a dataset description, including illumination, classes, and guidelines to generate prompts. We adapted it by replacing class descriptions with MALS attributes. The second LLM-generated prompt follows ALIA \cite{method_alia}, where captions of 20 random images are extracted from MALS and fed into the LLM to get additional variations. Dataset-derived prompts are based on MALS captions, and we consider two options: with and without prompt alignment, where alignment ensures that prompted attributes appear in the generated images. \cref{fig:fidcomprompts} shows the obtained results, where aligned captions perform better for every configuration, as it is easier to match attributes between the prompt and the init images.

\subsection{Results for blurring and context of images}
\label{subsubsec:img2imgimg}

\begin{figure}
\begin{center}
    \includegraphics[scale=0.35]{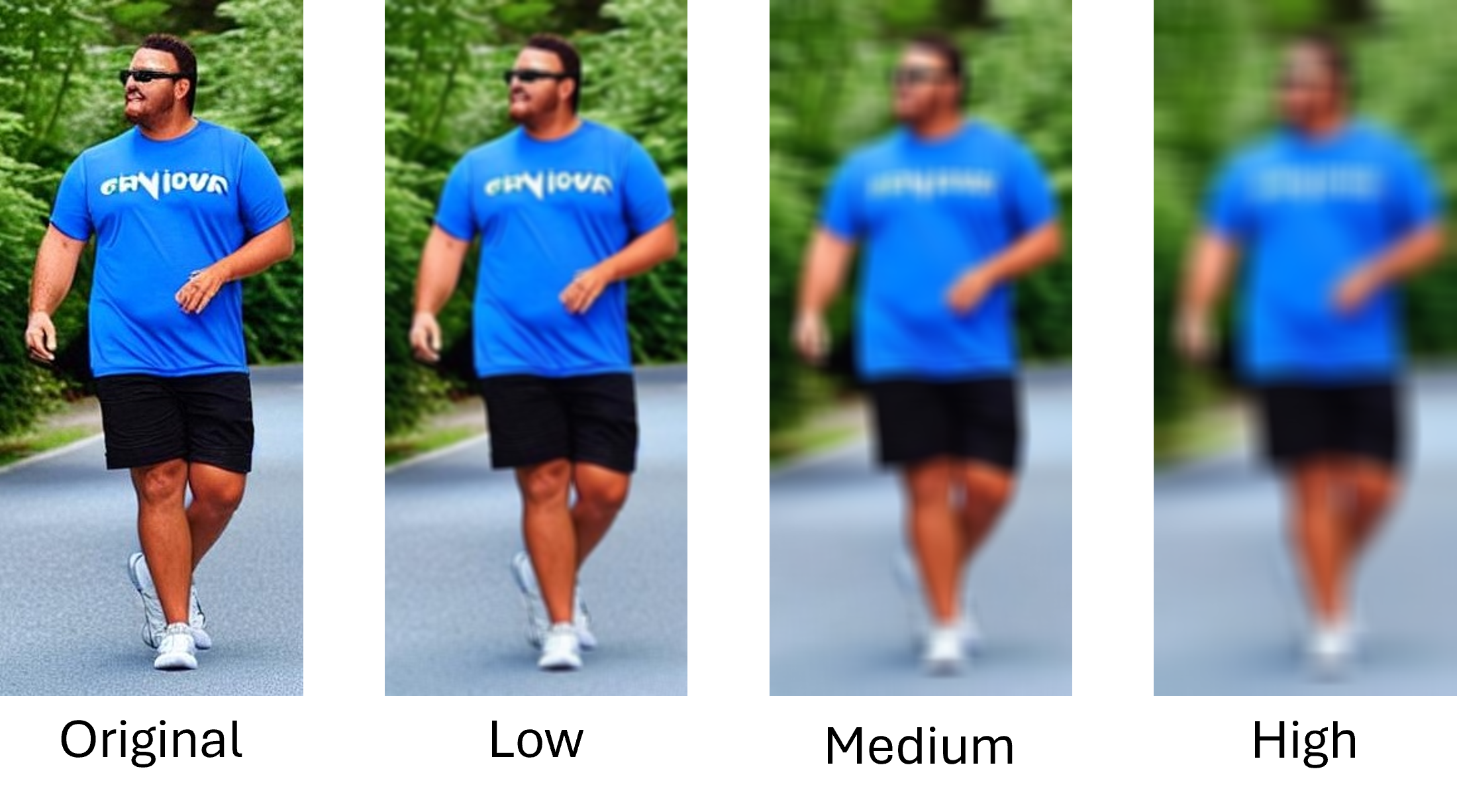}
    \caption{Examples of conditional images for blurring experiments. We applied Gaussian noise using three levels: low (kernel 5×5, standard deviation 5×5), medium (kernel 15×15, standard deviation 25×25) and high (kernel 25×25, standard deviation 50×50).}
    \label{fig:blurrexample}
    \vspace{-2mm}
\end{center}
\end{figure}

 To investigate the impact of \textbf{blurring}, we explored how applying different grades of noise to the conditional images will affect to the generation process. Given the results in  \cref{subsec:img2imgprompt}, we used the aligned MALS prompt. Where \cref{fig:blurrexample} shows an example of distortion grade of blurring levels explored. The results in \cref{fig:blurresults} shows that the more noise introduced, the higher the FID score in every configuration. Thus, the lack of clarity from RAPzs could be affected when generating images from it.

\begin{table}[]
\centering
\resizebox{0.7\columnwidth}{!}{
\begin{tabular}{l|ccc|}
\cline{2-4}
 & \multicolumn{3}{c|}{\textbf{FID score per configuration}} \\ \hline
\multicolumn{1}{|l|}{\textbf{Blurring}} & \multicolumn{1}{l|}{HiStr LoSc} & \multicolumn{1}{l|}{HiStr HiSc} & \multicolumn{1}{l|}{LoStr LoSc} \\ \hline
\multicolumn{1}{|l|}{Without} & \multicolumn{1}{c|}{\textbf{179}} & \multicolumn{1}{c|}{\textbf{185}} & \textbf{135} \\ \hline
\multicolumn{1}{|l|}{Low} & \multicolumn{1}{c|}{195} & \multicolumn{1}{c|}{198} & 166 \\ \hline
\multicolumn{1}{|l|}{Medium} & \multicolumn{1}{c|}{239} & \multicolumn{1}{c|}{213} & 208 \\ \hline
\multicolumn{1}{|l|}{High} & \multicolumn{1}{c|}{264} & \multicolumn{1}{c|}{228} & 267 \\ \hline
\end{tabular}
}
\caption{Comparative results for image-based experiments with blurring to generate PAR-like images. Baseline in bold.}
    \label{fig:blurresults}
\end{table}

\begin{figure}
    \centering
    \includegraphics[scale=0.30]{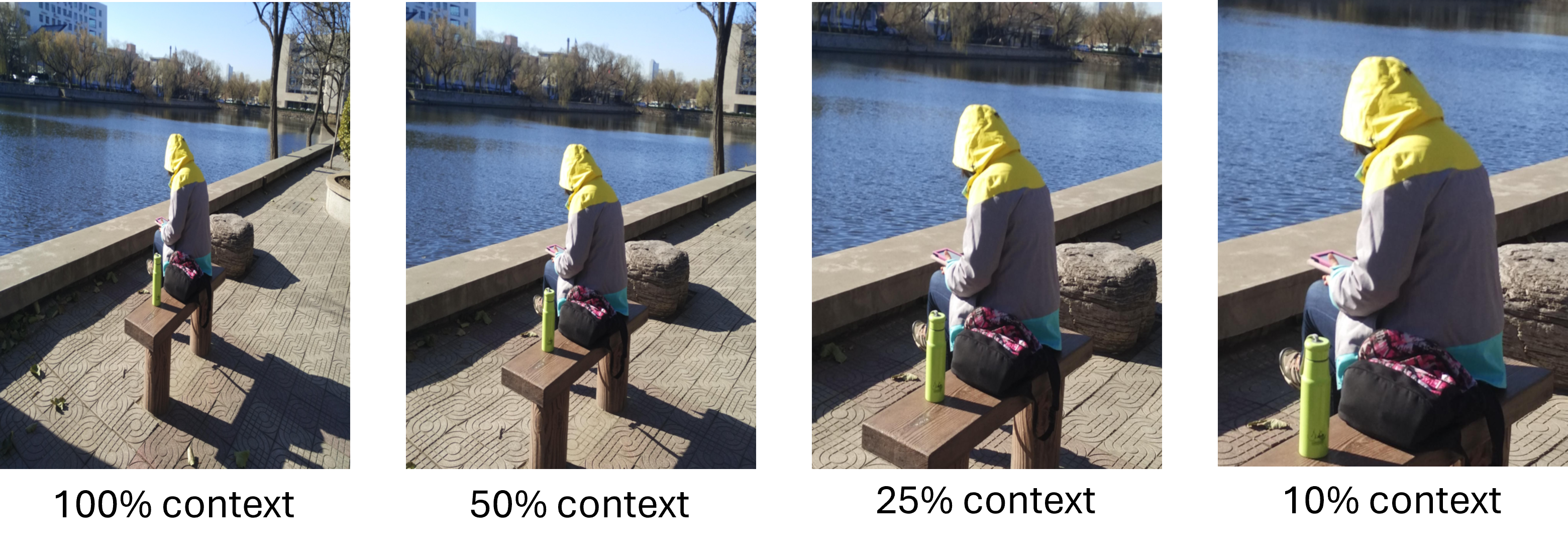}
    \caption{Examples of conditional images showing the pedestrian's bounding box with varying surrounding spatial context.}
    \label{fig:contextexample}
    \vspace{-2mm}
\end{figure}

Moreover, we analyzed influence of the \textbf{spatial context} of the pedestrian. \cref{fig:contextexample} shows examples for increasing contexts. Given the absence of PAR labels in the TJU-DHD dataset, we rely on MALS prompts (thus non-aligned). As shown in \cref{fig:contextresults}, the FID score is lower with LoSt \& LoSc and HiSt \& HiSc when the image has more context. Pedestrians occupy a larger area in the low context case (10\%) than in the high context case (100\%), resulting in a greater impact when context is reduced.

\begin{table}[]
\centering
\resizebox{0.6\columnwidth}{!}{
\begin{tabular}{c|ccc|}
\cline{2-4}
\multicolumn{1}{l|}{\textbf{}} & \multicolumn{3}{c|}{\textbf{FID score per configuration}} \\ \hline
\multicolumn{1}{|c|}{\textbf{Context}} & \multicolumn{1}{l|}{HiStr HiSc} & \multicolumn{1}{l|}{HiStr LoSc} & \multicolumn{1}{l|}{LoStr LoSc} \\ \hline
\multicolumn{1}{|c|}{100\%} & \multicolumn{1}{c|}{196} & \multicolumn{1}{c|}{213} & \textbf{85} \\ \hline
\multicolumn{1}{|c|}{50\%} & \multicolumn{1}{c|}{197} & \multicolumn{1}{c|}{206} & 86 \\ \hline
\multicolumn{1}{|c|}{25\%} & \multicolumn{1}{c|}{\textbf{195}} & \multicolumn{1}{c|}{\textbf{201}} & 94 \\ \hline
\multicolumn{1}{|c|}{10\%} & \multicolumn{1}{c|}{215} & \multicolumn{1}{c|}{210} & 106 \\ \hline
\end{tabular}

}
\caption{
Comparative results for context-based experiments to generate PAR-like images. Best results in bold. }
    \label{fig:contextresults}
\end{table}

\subsection{Results for diffusion-based techniques}
\label{subsec:img2imgmethod}

\begin{figure}[t]
\centering
\includegraphics[width=\columnwidth, height=0.2\textwidth]{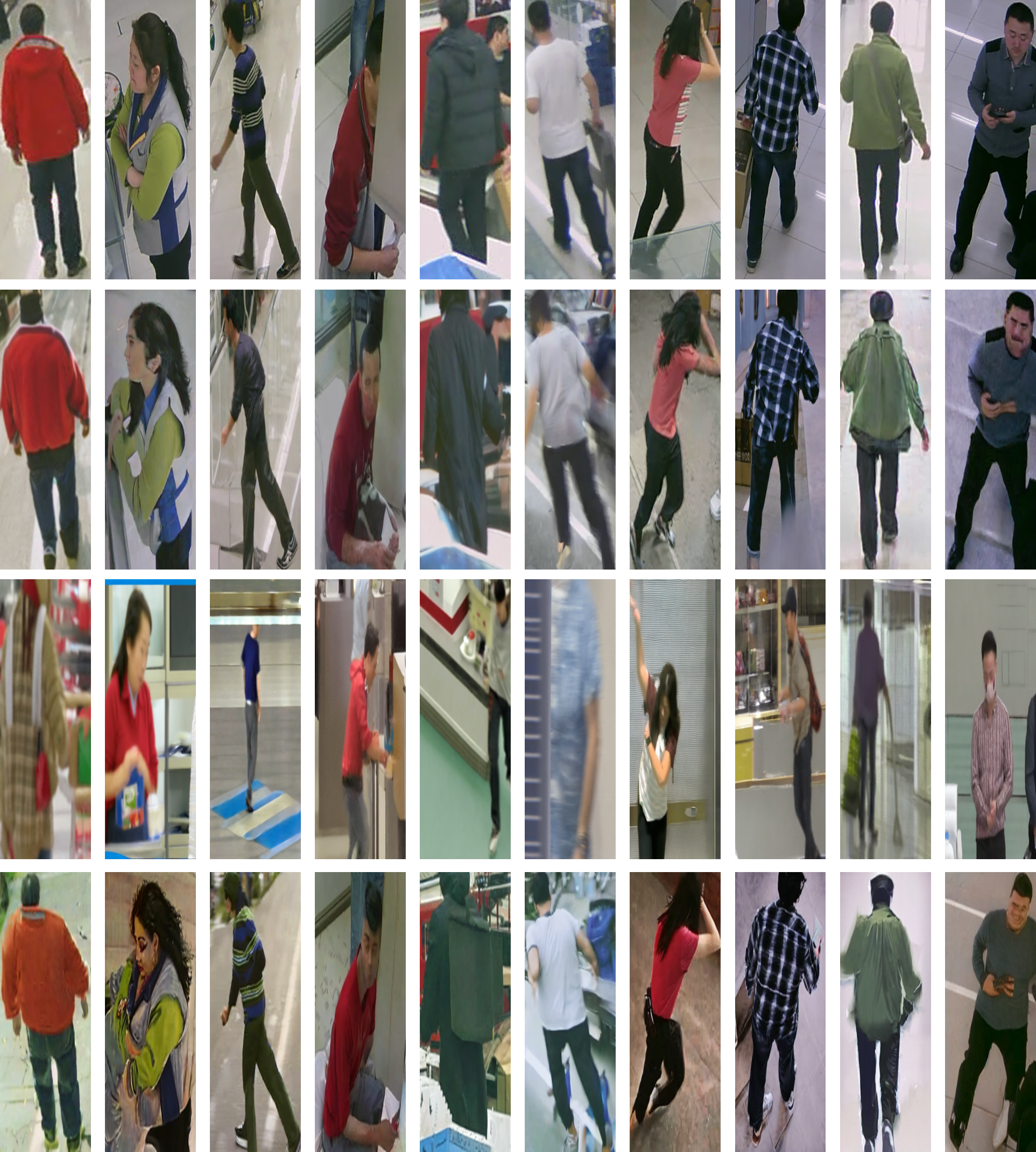}
\caption{Examples of generated images. First row: init img from RAPzs. Second row: DAFUSION \cite{method_dafusion}. Third row: DALDA \cite{method_dalda}. Fourth row: GIF \cite{method_gif}. All were obtained with HiSt HiSc config.} \label{figure:techniques_examples}
\vspace{-2mm}
\end{figure}

We also evaluated the performance of selected techniques in \cref{sec:canonicalsystem}. Here, we build a prompt for the RAPzs dataset with the following structure: [\textit{a/there is] a [gender] with [hair style] is [action] [attachment] wearing [upper-body] wearing [lower-body] wearing [type of shoe]}. Attributes are derived from available annotations from RAPzs \cite{rethinking_2021}. Hence, the prompt and the image are aligned (i.e. describing the same content) for RAPzs. Since each technique is class-based, we used "pedestrian" as class for all methods. Additionally, DAFUSION was trained using conditional images prior to the generation process. \cref{fig:tableTechnquies} shows that the best FID results are obtained with DAFUSION, indicating that the  Textual Inversion mechanism  is effective in the adaptation of the generated images to the target distribution. Also, in the \cref{figure:techniques_examples} we have a comparison between samples generated by each technique. Given FID and visual results, we selected DAFUSION for data expansion.

\begin{table}[]
\centering
\resizebox{\columnwidth}{!}{
\begin{tabular}{c|ccc|}
\cline{1-4}
\multicolumn{1}{|c|}{\textbf{Difusion-based Techniques}} & \multicolumn{1}{c|}{HiStr HiSc} & \multicolumn{1}{c|}{HiStr LoSc} & LoStr LoSc \\ \hline
\multicolumn{1}{|c|}{Textual Inversion (DAFUSION\cite{method_dafusion} )} & \multicolumn{1}{c|}{\textbf{164}} & \multicolumn{1}{c|}{\textbf{196}} & \textbf{135} \\ \hline
\multicolumn{1}{|c|}{Dynamic strength (DALDA \cite{method_dalda} )} & \multicolumn{1}{c|}{204} & \multicolumn{1}{c|}{209} & 203 \\ \hline
\multicolumn{1}{|c|}{Latent alteration (GIF\cite{method_gif} )} & \multicolumn{1}{c|}{190} & \multicolumn{1}{c|}{215} & 173 \\ \hline
\end{tabular}
}
\caption{FID comparative results for each configuration to generate PAR-like images. Best results in bold.}
    \label{fig:tableTechnquies}
\end{table}

\subsection{PARzs performance with dataset expansion}
\label{subsec:train_dafusion}

\subsubsection{PARzs dataset expansion}
\label{subsec:img2imgtotrain}

To expand the zero-shot datasets, we selected DAFUSION as it showed best results (see  \cref{subsec:img2imgmethod}). We changed Textual Inversion token learning from a global pedestrian class to per attribute classes (using subsets containing each attribute). Also, we used a prompt RAPzs and two adaptations for PETAzs and PA100k according to their attributes. Across all results in \cref{sec:expresults}, the LoStr-LoSc setup yields minimal variation between inputs, while HiSt-LoSc shows higher FID due to denoising-based guidance. Therefore, we chose the HiSt-HiSc configuration for data expansion. We used each dataset training samples as a conditioning image doubling the training set size. Post-generation FID scores are 44 (RAPzs), 34 (PA100K), and 33 (PETAzs), compared to reference train-test FIDs of 7, 13, and 12, respectively.

\subsubsection{PAR performance evaluation}

We trained a PAR method \cite{rethinking_2021} with the datasets expanded from \cref{subsec:img2imgtotrain}. Each synthetic sample is labeled only with attributes contained in the associated prompt. \cref{tab:results_sota} compares performance with popular data augmentations. We observe that training with the expanded dataset led to improvements over all methods and zero-shot datasets.

\begin{table}[]
\centering
\resizebox{0.8\columnwidth}{!}{
\begin{tabular}{|c|c|c|c|}
\hline
\textbf{Data Aug method} & \multicolumn{1}{l|}{\textbf{PA100k}} & \textbf{PETAzs} & \textbf{RAPzs} \\ \hline
Base \cite{rethinking_2021} & 80.44 & 71.47 & 71.98 \\ \hline
Autoaug \cite{autoaug} & 80.98 & 71.19 & 72.56 \\ \hline
CutMix \cite{cutmix} & 78.44 & 69.66 & 69.78 \\ \hline
Mixup \cite{mixup} & 77.83 & 69.13 & 69.41 \\ \hline
RandAug \cite{randaug} & 81.14 & 71.61 & 72.98 \\ \hline
Trivial \cite{trivaug} & 80.93 & 71.48 & 73.1 \\ \hline
Augmix \cite{augmix} & 80.73 & 71.71 & 73.02 \\ \hline
Ours & \textbf{81.58} & \textbf{73.07} & \textbf{75.23} \\ \hline
\end{tabular}
}
\caption{Comparative mA results against other data augmentation methods. Best results are highlighted in bold. All are trained using the same architecture \cite{rethinking_2021} and backbone (ResNet50 \cite{resnet50}).}
\label{tab:results_sota}
\vspace{-3mm}
\end{table}

%



\begin{figure}[t]
\centering
\includegraphics[scale=0.3]{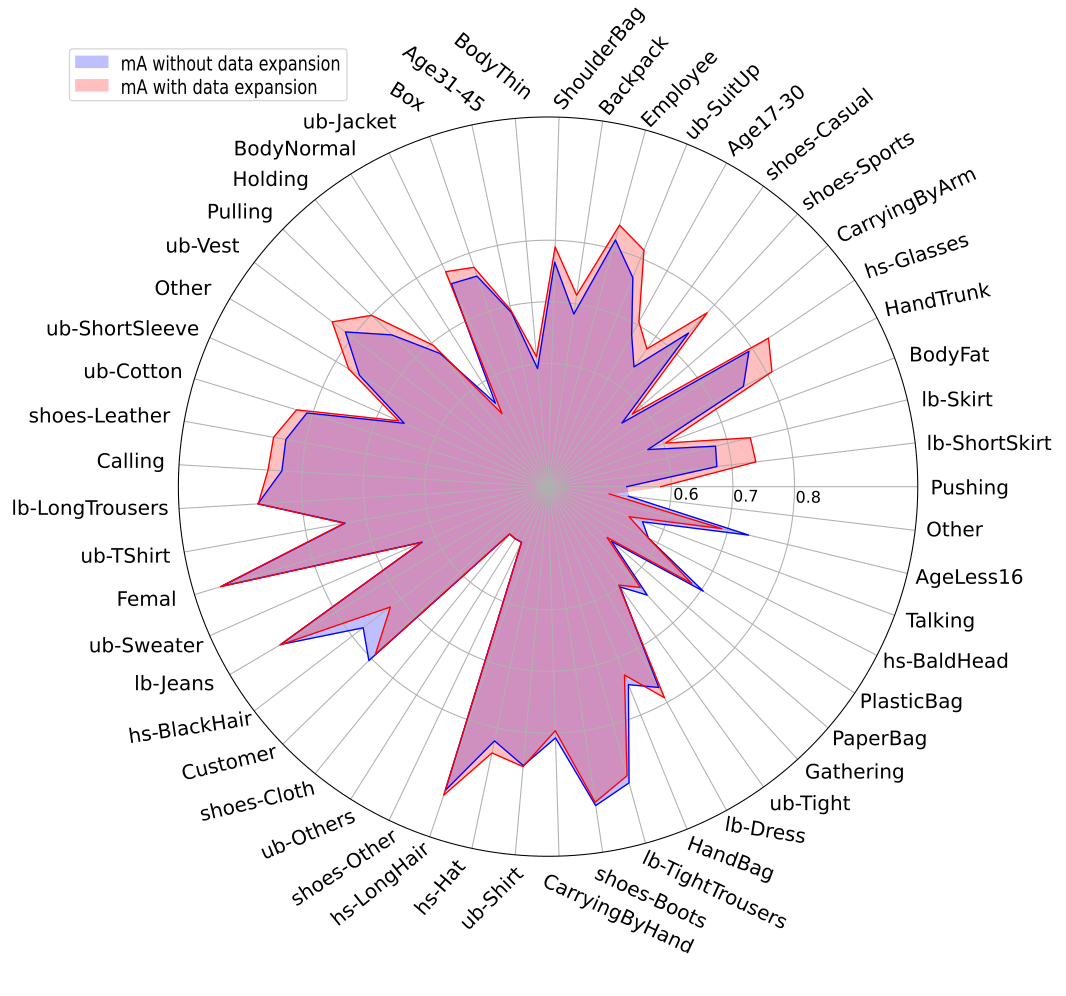}
\caption{Comparative results in terms of mA for each attribute of a PAR model \cite{rethinking_2021} training without and with synthetic data from DAFUSION \cite{method_dafusion} jointly with real data from the RAPzs dataset.} \label{figure:dafusion_training}
\vspace{-3mm}
\end{figure}

\cref{figure:dafusion_training} compares per-attribute test performance between expanded and baseline RAPzs datasets. We matched or improved the baseline performance, except for the attributes \textit{black hair} and \textit{ageLess16}. We enhanced underrepresented attributes with 200-900 training samples like \textit{handtruck}, \textit{backpack}, \textit{box}, \textit{shoulderbag}, \textit{skirt}, \textit{shortskirt}, \textit{hat}, \textit{calling}, \textit{pulling}, \textit{carrying by arm}, \textit{pushing}, \textit{suitup} or \textit{tight}. We also improved fine-grained attributes like \textit{hat}, \textit{glasses}, actions or type of clothe).

\section{Conclusions and future work}

This study examined the impact of prompt and image properties on the PAR-like image generation process. We explored four state-of-the-art techniques for producing images that closely resemble our target data. The FID scores between generated and original images highlighted a significant domain gap. Moreover, a PAR model trained on synthetic data achieved 1–4\% performance gains across multiple dataset as compared to popular augmentation strategies.

For future work, we aim to refine Textual Inversion for more precise attribute distribution learning, and expand our study to additional datasets and cutting-edge PAR methods.

\section*{Acknowledgement}
This work has been partially supported by the  Ministerio de Ciencia e Innovaci\'{o}n of the Spanish Government (grant PID2021-125051OB-I00) and by the Regional Government of Madrid of Spain (grant TEC 2024/COM-322).

{\small
\bibliographystyle{ieeetr}

\bibliography{egbib}
}

\clearpage
\appendix
\onecolumn
\begin{center}
\large{\bf{SUPPLEMENTARY MATERIAL}}    
\end{center}

This supplementary material is provided to support and enhance the findings presented in the main manuscript. It includes additional figures, tables and additional experiments that could not be accommodated in the main text. These materials aim to offer deeper insights, ensure transparency, and provide sufficient information for replication and further exploration of the study.

\section{Prompts adaptation and examples}
\label{sup_sect:prompt_explanation}

RAPzs and baseline prompts were already explained in the paper. Thus, we will explain the others prompts used. 

\subsection{PETAzs and PA100k prompt adaptation.}
\label{sup_sect:prompt_dafusion_multi}

PA100k and PETAzs are similar to RAPzs but adapted to their attributes. 
We created the PETAzs prompt with the following structure:

\textit{\textbf{\textit{Hand-crafted template}} (A, There is a)
\textbf{\textit{Gender}} (woman, man) with \textbf{\textit{Hair Color}} (black, blue, brown, green, grey, orange, pink, purple, red, white, yellow) \textbf{\textit{Hair}} (long hair, bald, short hair) wearing  \textbf{\textit{Upper body}} (casual, formal, jacket, logo, plaid, thin stripes, t shirt, other, vneck) \textbf{\textit{Color}} (black, blue, brown, green, grey, orange, pink, purple, red, white, yellow) \textbf{\textit{Lower body}} (casual, formal, jeans, shorts, shortskirt, trousers, capri, hotpants, long skirt, plaid, thin stripes, suits) \textbf{\textit{Color}} carrying \textbf{\textit{Attachment}} (backpack, other, messenger bag, nothing, plastic bags, baby buggy, shopping tro, umbrella, folder, luggage case, suit case) with \textbf{\textit{Accessory}} (hat, muffler, nothing, sunglasses, headphone, hairband, kerchief) wearing \textbf{\textit{Color}} \textbf{\textit{Footwear}} (leather shoes, sandals, shoes, sneakers, stocking).}

Here, there is an example of PETAzs prompt:

\textit{A man with black short hair carrying other with hat wearing black casual jacket logo grey casual shorts grey sneakers.}
\newline

Also, we created the PA100k prompt with the following structure: 

\textit{\textbf{\textit{Hand-crafted template}} (A, There is a)
\textbf{\textit{Gender}} (woman, man) from \textbf{\textit{View}} (front, side, back) with \textbf{\textit{Attachment}} (hand bag, shoulder bag, back pack, holding object in front) wearing \textbf{\textit{Upper Wearing}} (short sleeve, long sleeve, long coat) \textbf{\textit{Logo}} (logo, plaid, splice) \textbf{\textit{Lower Clothes}} (trousers, short, skirt or dress) \textbf{\textit{Lower Type Clothes}} (stripe, pattern) \textbf{\textit{shoes}} (boots).}

Being an example of PA100k prompt:

\textit{There is a woman from side with shoulder bag wearing long sleeve trousers.}
\newline
\newline
After the generation, only the attributes from \citesupp{rethinking_2021} were taked into account. For PA100k attributes we selected: Gender, View, Attachment, Lower Clothes, Lower Type Clothes and boots. In the case of PETAzs attributes we selected: Gender, Hair, Upper Body, Lowe Body, Accessory, Footwear.

\subsection{Prompt integration}
\label{sup_sect:prompt_integration}

The prompt integration \citesupp{prompt_integrating} has the following structure: 

\textit{\textbf{\textit{Hand-crafted template}} (There is a, A photo of)
\textbf{\textit{Articles}} (a, an, the) \textbf{\textit{Ages}} (young, old, little, elderly) \textbf{\textit{Bodies}} (tall, short, big, small) \textbf{\textit{Expressions}} (smiling, crying, displeased) \textbf{\textit{Classes}} (pedestrian) \textbf{\textit{articles clothes}} (in, wearing, with) \textbf{\textit{clothes}} (t-shirt, dress, jeans, hat, hair) \textbf{\textit{colors}} (white, black, red, blue, yellow, orange) \textbf{\textit{poses}} (standing, walking, sitting, crouching) \textbf{\textit{directions}} (in front, in profile, from behind) \textbf{\textit{actions}} (walking).}

Here there is an example of a prompt:

\textit{A photo of an elderly tall crying pedestrian wearing sweater red sitting in front.}

\subsection{Prompt MALS}
\label{sup_sect:prompt_mals}

Since MALS \citesupp{mals_dataset} prompts are synthetic generated do not have an strict structure but here we have a default one: 

\textit{A \textbf{\textit{subject}} in a \textbf{\textit{clothing description}} \textbf{\textit{action}} with \textbf{\textit{object description}} in \textbf{\textit{setting}}, \textbf{\textit{optional detail}}.}

Being an example: 

\textit{A man walking down a sidewalk with a bag in his hand and a handbag in his other hand, with trees in the background.}

\subsection{Prompt DALDA}
\label{sup_sect:prompt_dalda}

The DALDA \citesupp{method_dalda} prompt are generated by GPT-4. They gives a data-set description, a list of classes and some instructions. GPT-4 will create one prompt for each class, this is repetead to have several prompts for each class. In our case, the prompts were adapted to multi-attribute, we have changed the prompt for GPT-4 to have different prompts. Being the prompt requested to the LLM: 

\textit{RAPzs dataset are images of pedestrians with attributes. The images are captured by a video surveillance cameras at indoor scenes. Each image has a caption where attributes are described. The size of each image is around 90 x 180 pixels. The attribute categories will be explained as 'attribute category': 'option1, option2..'. 
The attribute categories are the following, in each attribute category you will find each options that you can select for the prompt, the binary ones does not have options, only to put the category at the prompt or not:
'head' [BaldHead, LongHair, BlackHair, Hat, Glasses], 'upper body' [Shirt, Sweater, Vest, TShirt, Cotton, Jacket, SuitUp, Tight, ShortSleeve, Others], 'lower body' [LongTrousers, Skirt, ShortSkirt, Dress, Jeans, TightTrousers], 'footwear' [shoes-Leather, "shoes-Sports, shoes-Boots, shoes-Cloth, shoes-Casual, shoes-Other], 'accessory" [attachment-Backpack, attachment-ShoulderBag, attachment-HandBag, attachment-Box, attachment-PlasticBag, attachment-PaperBag, attachment-HandTrunk, attachment-Other], 'age' [AgeLess16, Age17-30, Age31-45, Age46-60], 'gender' [Male, Female], 'body shape' [BodyFat, BodyNormal, BodyThin], 'role' [Customer, Employee], 'action' [action-Calling, action-Talking, action-Gathering, action-Holding, action-Pushing, action-Pulling, action-CarryingByArm, action-CarryingByHand, action-Other].
\begin{itemize}
    \item You must create a sentence that describes the attribute-oriented image corresponding to the 'dataset description' and a combination of each attribute category and their options.
    \item While maintaining the properties of each attribute category, we will add external elements other than the class to the statement.
    \item The attributes added to each sentence must fit the 'dataset description' and be realistic. 
    \item Find the best combination sentence of the category attributes to generate images to image.
\end{itemize}
So you must create 100 sentence with a combination of each attribute category, remind that the binary categories should be used with their name or not appear at the sentence.
The output should always be in json format, no text other than json is required like this: 'prompt1': 'sentence', 'prompt2': 'sentence' }
\newline

As a result, the LLM gave prompts like this: 
\textit{A bald male with a tight shirt and long trousers, wearing shoes-Leather, carrying attachment-Backpack, in an indoor shopping mall.}

\subsection{Prompt ALIA}
\label{sup_sect:prompt_alia}

The ALIA \citesupp{method_alia} prompt are also created with GPT-4 or Vicuna. But instead of only using a prompt, they used BLIP to get all the captions from the data-set, select 20 random captions and create a prompt to GPT-4. They tries to generate variations with GPT-4 to introduce more variability. Thus, the prompt used to the LLM is: 

"I have a set of image captions that I want to summarize into objective descriptions that describe the scenes, actions, camera pose, zoom, and other image qualities present. 
My captions are: 
{captions}
I want the output to be a less than 10 of captions that describe a unique setting, of the form \"{prefix}\".
Here are 1 examples of what I want the output to look like:
- A male with a hat, in a jacket and long trousers, shoes-Casual, carrying attachment-PlasticBag, walking through a gallery"

An example of the prompt given by ALIA: 

\textit{A woman in a winter coat and boots, carrying attachment-Bag, standing in snow near a scenic view.}

\section{Additional experiments}

\subsection{Image to image based: resolution based}

\begin{figure}[H]
\centering
\includegraphics[scale=0.7]{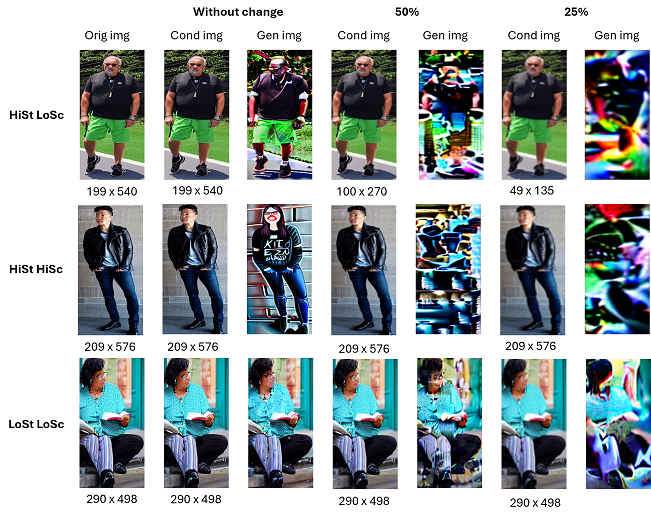}
\caption{Visual examples of resolution experiments.}
\label{fig:sup_expresolution}
\end{figure}

In this case, we have analysed the impact
of reducing the resolution from the image (maintaining
the aspect). Thus, we use images from the MALS dataset (high resolution for example 159x570) and reduce their resolution by 50\% and 25\%. We can see an example of the resolution reduction in the \cref{fig:sup_expresolution}. In the \cref{tab:sup_expresolution} we can see that reducing the resolution of the image the generated image the FID increased. Thus, with the visual results and the generated images we can conclude that reducing the resolution of the input image highly impacts in the generation process.

\begin{table}[H]
\centering
\begin{tabular}{|c|c|c|c|}
\hline
\textbf{Resolution} & \textbf{HiSt HiSc} & \textbf{HiSt LoSc} & \textbf{LoSt LoSc} \\ \hline
\textbf{25\%} & 399 & 397 & 316 \\ \hline
\textbf{50\%} & 356 & 342 & 210 \\ \hline
\end{tabular}
\caption{Results for resolution experiments. Compared against MALS dataset.}
\label{tab:sup_expresolution}
\end{table}

\subsection{Image to image based: aspect ratio based}

\begin{figure}[t]
\centering
\includegraphics[scale=0.62]{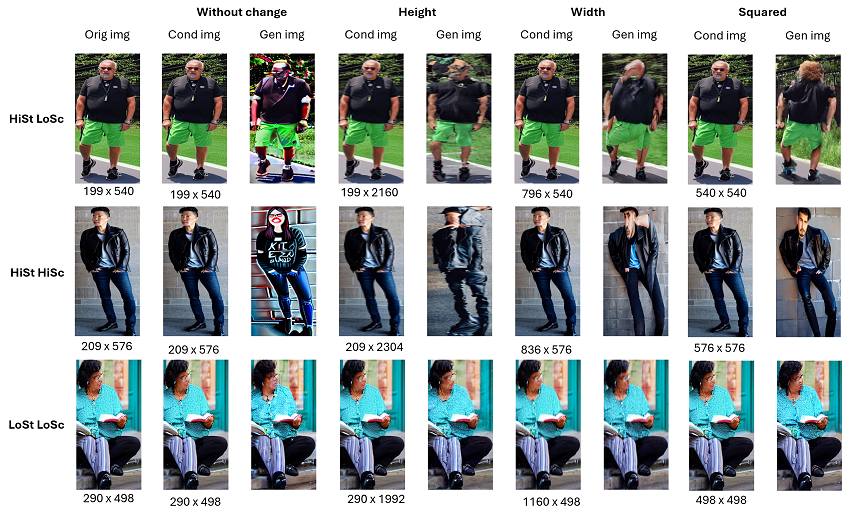}
\caption{Example of aspect ratio experiments.}
\label{fig:sup_expaspect}
\end{figure}

In this experiment we explored how the
image aspect ratio affects to the generation, with a square, long width and long height aspect. In the case of squared, we set the width same as height. In the \cref{fig:sup_expaspect} we can see examples of the original image, the changed image used as conditional image and the generated image for the different configuration of strength and scale. Also, in the \cref{tab:sup_expaspect} we can see how the aspect ratio does not affect in the case of LoSt and LoSc but that is because the conditioning is insignificant. In the other configurations, we can see that it is better to have a width longer than squared or too long from height.

\begin{table}[t]
\centering
\begin{tabular}{|c|c|c|c|}
\hline
\textbf{Aspect} & \textbf{HiSt HiSc} & \textbf{HiSt LoSc} & \textbf{LoSt LoSc} \\ \hline
\textbf{Squared} & 147 & 176 & 120 \\ \hline
\textbf{Width} & 150 & 154 & 121 \\ \hline
\textbf{Height} & 205 & 174 & 124 \\ \hline
\end{tabular}
\caption{Aspect ratio experiments with FID table compared against MALS dataset. 100 images generated per relationship aspect.}
\label{tab:sup_expaspect}
\vspace{0mm}
\end{table}

\section{Additional visual results from experiments}
\subsection{Image to image based: blurring based}
\begin{figure}[t]
\centering
\includegraphics[scale=0.6]{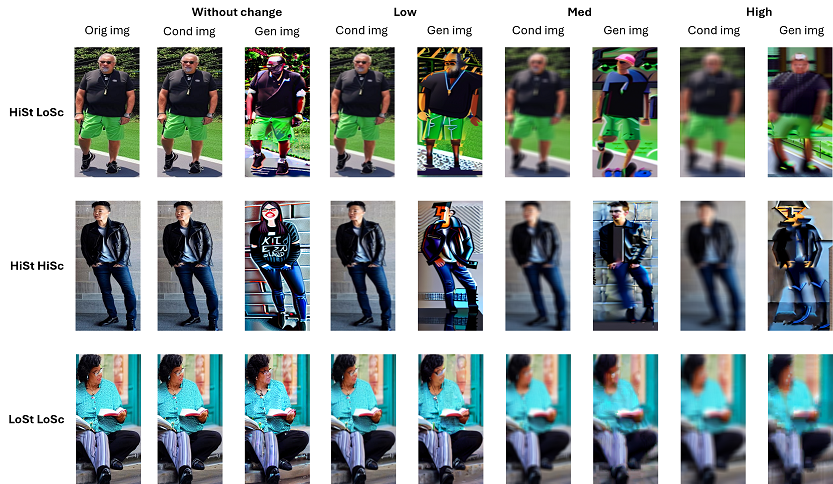}
\caption{Visual results of blurring experiments.}
\label{sup_mat:fig_blurring}
\vspace{-4mm}
\end{figure}

In this subsection we wanted to show visual results from the blurring experiment presented in the main document. In the \cref{sup_mat:fig_blurring} we can check how applying different blurring to the conditional image does not affect at all since without blurring the generated images are different from the original distribution, but the shape of the person is maintained. Also, in the LoSt and LoSc the difference is very small.

\subsection{Image to image based: context based}

\begin{figure}[b]
\centering
\includegraphics[scale=0.6]{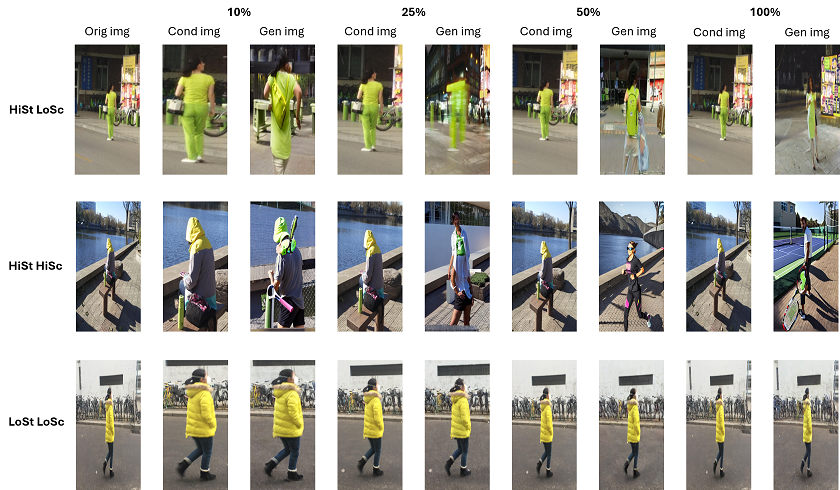}
\caption{Visual results of context experiments.}
\label{sup_mat:fig_context}
\end{figure}

In this subsection we wanted to show visual results from the context experiment presented in the main document. In the \cref{sup_mat:fig_context} we can see how adding context to the conditioning image could affect in the HiSt and HiSc configuration since the conditioning is higher and changes more original image. But in the configuration of HiSt and LoSc the images generated are similar to the original but making data augmentation. As in every experiment with the LoSt and LoSc configuration, the difference of the generated image is very small.

\subsection{Dataset expansion examples: DAFUSION multiattribute}

\begin{figure}[]
\centering
\includegraphics[scale=0.75]{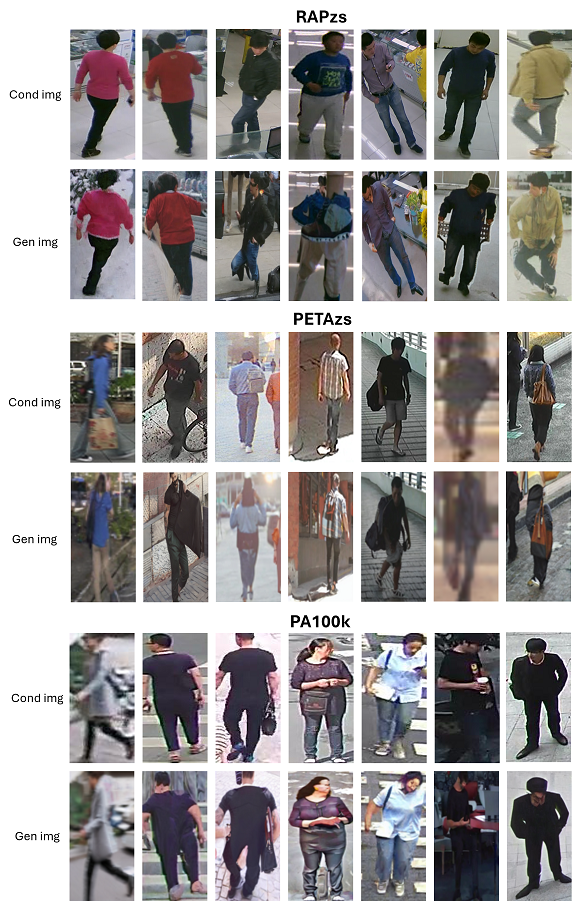}
\caption{Visual examples of dataset expansion with each dataset explored. For RAPzs \cite{rethinking_2021}, PETAzs \cite{rethinking_2021} and PA100k \cite{pa100k_dataset}, first row conditional image and at the second row the generated image.}
\label{sup_mat:fig_datasetexpaned}
\end{figure}

In this subsection we show visual results from the datasets expanded. In the \cref{sup_mat:fig_datasetexpaned} we can see examples of images from the datasets explored and the generated images with DAFUSION \citesupp{method_dafusion} trained as multiattribute and with the corresponding dataset prompt.

\section{Additional setup}
\subsection{Generation}

In the prompt and image based experiments we used
\href{https://huggingface.co/CompVis/stable-diffusion-v1-4}{Stable diffusion} from HuggingFaces. Also, we used \href{https://github.com/kkyuhun94/dalda}{DALDA} and \href{https://github.com/lisadunlap/ALIA}{ALIA}. Then, for the techniques methods we used their repositories: \href{https://github.com/kkyuhun94/dalda}{DALDA}, \href{https://github.com/brandontrabucco/da-fusion}{DA-Fusion}, \href{https://github.com/Vanint/DatasetExpansion}{GIF}.

\subsection{Training PAR}
We used  In PAR training, we used the following \href{https://github.com/valencebond/Rethinking_of_PAR}{code}. Thus, we used a Resnet50 as a feature extractor, and then, a linear classifier to classify the attributes from features. The linear classifier is the only part that is trained. We used the SGD optimizer with 0.9 of momentum and a weigth decay of 0.0001. Moreover, the lr scheduler is plateu with LR\_FR, LR\_NEW of 0.01 and WMUP\_COEF of 0.1. The loss function is the bce. The image input are resized to 256x192 (height x width). 

\newpage

{\small
\bibliographystylesupp{ieeetr}
\bibliographysupp{egbib_supp}
}

\end{document}